\documentclass[10pt, conference, compsocconf]{IEEEtran}
\usepackage[ruled,vlined,linesnumbered]{algorithm2e}
\usepackage{color,graphicx,epstopdf,changepage,amsmath,multirow,amssymb,mathabx}
\usepackage[justification=centering]{caption}
\usepackage{blindtext}
\let\labelindent\relax
\usepackage[inline]{enumitem}
\usepackage{dblfloatfix}    
\usepackage{xcolor}
\usepackage{subcaption}
\SetAlgoCaptionSeparator{.\space}

\usepackage{flushend}
\usepackage[pdfauthor={\@author}, pdftitle={\@title}]{hyperref}
\makeatother

\vspace{-1.5cm}
\makeatletter
\g@addto@macro\normalsize{%
  \setlength\abovedisplayskip{3pt}
  \setlength\belowdisplayskip{3pt}
  \setlength\abovedisplayshortskip{3pt}
  \setlength\belowdisplayshortskip{3pt}
  \setlength{\textfloatsep}{3pt plus 1.0pt minus 2.0pt}
  \setlength{\intextsep}{3pt}
}
\makeatother

\newcommand{\gra}{\mathcal{G}}

\renewcommand{\mathbf}[1]{{\color{green}{#1}}}

\newcommand{\textloss}[1]{\textcolor{red}{\bf #1}}

\newcommand{\pop}[1]{\mathcal P^{#1}}

\begin{document}
%
\title{Evolutionary Multitasking for Semantic Web Service Composition}

	\author{
	\IEEEauthorblockN{Chen Wang, Hui Ma, Gang Chen}
		\IEEEauthorblockA{School of Engineering and Computer Science\\ 
		Victoria University of Wellington\\
		New Zealand\\
		Email: \{chen.wang, hui.ma, aaron.chen\}@ecs.vuw.ac.nz}
	\and
	\IEEEauthorblockN{Sven Hartmann}
		\IEEEauthorblockA{Department of Inkformatics\\ 
		Clausthal University of Technology\\
		Germany\\
		Email: sven.hartmann@tu-clausthal.de}

	}

\maketitle


\begin{abstract}
Web services are basic functions of a software system to support the concept of service-oriented architecture. They are often composed together to provide added values,  known as web service composition. Researchers often employ Evolutionary Computation techniques to efficiently construct composite services with near-optimized functional quality (i.e., Quality of Semantic Matchmaking) or non-functional quality (i.e., Quality of Service) or both due to the complexity of this problem. With a significant increase in service composition requests, many composition requests have similar input and output requirements but may vary due to different preferences from different user segments. This problem is often treated as a multi-objective service composition so as to cope with different preferences from different user segments simultaneously. Without taking a multi-objective approach that gives rise to a solution selection challenge, we perceive multiple similar service composition requests as jointly forming an evolutionary multi-tasking problem in this work. We propose an effective permutation-based evolutionary multi-tasking approach that can simultaneously generate a set of solutions, with one for each service request. We also introduce a neighborhood structure over multiple tasks to allow newly evolved solutions to be evaluated on related tasks. Our proposed method can perform better at the cost of only a fraction of time, compared to one state-of-art single-tasking EC-based method. We also found that the use of the proper neighborhood structure can enhance the effectiveness of our approach.
\end{abstract}


%
\IEEEpeerreviewmaketitle

\section{Introduction}\label{section:Introduction}
\textit{Web service Composition} has been widely adopted in web service based systems as a computing paradigm for rapidly building up cost-efficient and integratable enterprise applications \cite{curbera2001web}. This composition is achieved by loosely coupling web services into execution workflows to provide added values for service users. Since these workflows are often unknown or not given by users, many researchers have been working on automatically constructing composition workflows with an aim to optimizing the overall quality of composite services \cite{wang2017comprehensive,wang2017gp,wang2018towards,wang2018eda,wang2018knowledge}.  The overall quality refers to Quality of Semantic Matchmaking  (QoSM) and Quality of Service (QoS) that are simultaneously optimized for functional and non-functional attributes of composite service respectively \cite{wang2017comprehensive}. 

Service composition problem is NP-hard and cannot be solved in polynomial time \cite{rao2005survey}. Due to this reason, Evolutionary computation (EC) techniques have been proposed to efficiently find near-optimal solutions that satisfy users' requirements reasonably well \cite{wang2017comprehensive,wang2017gp,wang2018towards,wang2018knowledge,rodriguez2010composition,da2016genetic,da2018evolutionary,yu2013adaptive,da2018hybrid,da2017fragment}. These EC-based service composition approaches are mainly classified into two groups based on the number of objectives to be optimized: single-objective \cite{wang2017comprehensive,wang2017gp,wang2018towards,wang2018knowledge,rodriguez2010composition,da2016genetic,da2018evolutionary,yu2013adaptive} or multi-objective  approaches \cite{da2018hybrid,da2017fragment}. The first group optimizes only one objective by combining all quality criteria into one (e.g., one combined quality that measures QoSM and QoS \cite{wang2017comprehensive}); the second group has an aim to identify a group of composite services with varied trade-offs over multiple objectives (e.g., two trade-off objectives:  one combines time and cost, the other combines availability and reliability \cite{da2018hybrid}).

Recently,  Gupta et al.~\cite{gupta2016multifactorial} introduced a new EC computing paradigm, namely, multifactorial evolutionary algorithm (MFEA)~\cite{gupta2016multifactorial}  with a unified random-key representation to search solutions for multiple tasks (or optimization problems) simultaneously. MFEA transfers implicit knowledge of promising solutions through the use of simple genetic operators across multiple tasks. These genetic operators allow two randomly selected parents to undergo crossover or mutation with certain conditions on the tasks. This genetic mechanism is called \textit{assortative mating} \cite{gupta2016multifactorial}. Besides, the offspring is only evaluated on one selected task determined by its parents based on vertical cultural transmission. MFEA has shown its efficiency and effectiveness in several problem domains~\cite{bao2018evolutionary,feng2015memes,yuan2016evolutionary,zhou2016evolutionary}.

Existing service composition algorithms are designed primarily to solve each service composition request independently by using either single-objective \cite{wang2017gp,rodriguez2010composition,da2016genetic,yu2013adaptive} or multi-objective approaches~\cite{da2018hybrid,da2017fragment}, ignoring similarities between different requests that could be dealt with collectively.  For example, many service composition requests have similar input and output requirements but may vary due to different preferences on their quality. These requests are handled repetitively without meeting efficiency and time requirements. 

In a market-oriented environment, a service composer often strategically groups all the users (i.e., service requesters) into several segments, e.g., platinum, gold, silver and bronze user segments. Composition tasks for users in one segment are packaged as one service composition task according to users' preferences. One segment offers (i.e., one published composite service) will serve each user in this segment separately and uniquely. Therefore, before computing a composition solution for any incoming request from scratch (which is expensive) we will check whether we have solved a similar request, and we can reuse the segment offer. For example, TripPlanner (an imaginary composite service used by travel agencies) provides travel planning support by coupling many external existing services, such as airline booking, hotel reservation, and payment services. The platinum segment (e.g., large international travel companies) has high QoSM requirements as their loyal customers often demand reliable and accurate information. In contrast, the bronze segment (e.g., small local travel companies), may care more about service cost than QoSM. Therefore, we can provide one segment offer to any incoming request for TripPlanner immediately by identifying the respective user segment.

The problem discussed above could be treated as multi-objective service composition so as to cope with different preferences from different user segments, though this is likely to result in a solution selection challenge. As an alternative, we perceive multiple similar service composition requests from segments as jointly forming an evolutionary multi-tasking problem in this work. Our goal is to evolve simultaneously a set of composition solutions, one for each composition task.  

Very recently, \cite{bao2018evolutionary} reported the first attempt to search optimal solutions for two unrelated composition tasks concurrently using MFEA, outperforming some basic single-objective EC techniques.
Despite this recent success, this work~\cite{bao2018evolutionary} can only handle semi-automated service composition problems, i.e., a specific service workflow must be given in advance and strictly obeyed. QoSM with segment preferences is not studied at all. Besides that,  the number of composition tasks that are optimized concurrently is very small (e.g.,  2 tasks in~\cite{bao2018evolutionary}), and the test cases used for the experiments are small (e.g., each test case only contains 2507 atomic web services in~\cite{bao2018evolutionary}). Furthermore, the findings of their experiments are based on comparisons with some basic EC techniques, overlooking state-of-art service composition approaches. To address these limitations above, we propose a novel \textbf{P}ermutation-based \textbf{M}ultifactorial \textbf{E}volutionary \textbf{A}lgorithm (henceforth referred to as \text{PMFEA}) to solve the fully automated semantic service composition problem for diverse user segments with different QoSM preferences. We aim to optimize the overall quality of the composition solutions (i.e., QoSM and QoS). The contributions of this paper are as follows:

\begin{enumerate}
\item We model multiple service requests for diverse user segments with different QoSM preferences as a multi-tasking problem. This is the first time in literature to formulate such a multi-tasking and fully automated service composition problem. We also propose PMFEA  to effectively and efficiently handle this problem with a permutation-based representation, using crossover and mutation \cite{da2018evolutionary} to support our presentation in assortative mating.

\item We further introduce a neighborhood structure over multiple tasks to allow newly evolved solutions to be additionally evaluated on the neighboring tasks. The use of this neighborhood structure has a severe impact on the effectiveness as well as the efficiency of PMFEA for optimizing more than two tasks concurrently.

\item We conduct experiments to explore the performance of PMFEA and two of its variations: PMFEA with evaluations on  \textbf{N}eighboring \textbf{T}asks (henceforth referred to as PMFEA-NT), and PMFEA with evaluations on \textbf{A}ll \textbf{T}asks (henceforth referred to as PMFEA-AT) and to compare them against a state-of-the-art fully automated service composition approach \cite{da2018evolutionary}. For the experiments we use a large benchmark dataset with multiple test cases of different sizes. The evaluation shows that all PMFEA approaches perform better at the cost of only a fraction of time. In particular, PMFEA-NT achieves the best performance in terms of effectiveness and efficiency.
\end{enumerate}
\section{Related Work}\label{section:relatedWork}
EC has been widely used in fully automated service composition problems to find near-optimal solutions efficiently, where QoSM or QoS, or both are optimized \cite{wang2017comprehensive,wang2017gp,wang2018towards,wang2018knowledge,rodriguez2010composition,da2016genetic,da2018evolutionary,da2018hybrid,da2017fragment,bao2018evolutionary}.  Based on the number of tasks to be optimized, those approaches can be classified into two main groups: evolutionary single-tasking and multi-tasking.

Evolutionary single-tasking approaches can be divided into two subgroups: single-tasking single-objective and single-tasking multi-objective approaches based on the number of objectives to be optimized.  The first subgroup is well studied with algorithm-dependent representations and breeding operators. Genetic programming has been employed to evolve tree-based composite solutions \cite{wang2017gp,rodriguez2010composition,da2016genetic,yu2013adaptive}. For example, a recent GP-based approach \cite{wang2017gp} proposed a tree-like representation for composite services. This representation allows better scalability by eliminating the replicas of subtrees based on the tree-based representations in \cite{rodriguez2010composition,da2016genetic,yu2013adaptive}. Apart from evolving trees, graph evolution techniques are employed to evolve graph-based composite solutions, such as GraphEvol~\cite{da2015graphevol}. On the other hand, Particle Swarm Optimization, Genetic Algorithm, and Estimation of Distribution Algorithm have been employed to solve the single-tasking service composition problem~\cite{wang2017comprehensive,wang2018towards,wang2018knowledge,da2018evolutionary}. For example, \cite{wang2018knowledge} samples permutation-based solutions from learning the distribution models of promising solutions in each generation. The second subgroup only reported two recent attempts on single-tasking multi-objective fully automated web service composition  \cite{da2018hybrid,da2017fragment}. For example, a hybrid approach \cite{da2018hybrid} was proposed to decompose the multi-objective problem into single-objective subproblems, producing a set of Pareto solutions with trade-offs. In practice, an indispensable decision must be made to choose a small number of solutions, which satisfy users' preferences. When there are many solutions in the Pareto front for more than two objectives, selecting a solution is a challenging task because users have to evaluate the trade-offs among the compositions manually.

Evolutionary multi-tasking is a new optimization paradigm, which has been proposed to solve combinatorial optimization problems \cite{bao2018evolutionary,feng2015memes,yuan2016evolutionary,zhou2016evolutionary} and produces multiple solutions, with one for each task. For example, Yuan et al. \cite{yuan2016evolutionary} employed MFEA to concurrently handle four optimization problems with different local search operators. The success of this work is attributed to the use of permutation-based representations and the expensive local search. Compared to the default random-key representation in MFEA, permutation-based representations are often effective for dealing with permutation-based problems, such as TSP, QAP, LOP, and JSP \cite{yuan2016evolutionary}. It is due to that decoding random-key representation to permutations is inefficient and can be highly lossy since only information on the relative order is derived \cite{yuan2016evolutionary}.  Apart from that,  existing studies in evolutionary multitasking have not considered a neighborhood structure over tasks (or optimization problems) to allow newly evolved solutions to be evaluated on related tasks. In this paper, we will employ a permutation-based representation and introduce a neighborhood structure over a set of composition tasks. It should be noted that very few service composition works employ evolutionary multi-tasking. To the best of our knowledge, \cite{bao2018evolutionary} reported the first attempt to optimize just two composition tasks concurrently for semi-automated service composition. We have addressed the limitations of that work in Section~\ref{section:Introduction} above. In particular,  it cannot handle the fully automated service composition problem studied in this paper. The motivation for our research is to overcome these limitations.
\section{Preliminaries}\label{section:Preliminary}
In this section, we first present the concepts of multifactorial optimization. We then formulate our web service composition problem as a multitasking problem.
\subsection{Multifactorial Optimization}\label{section:prooblem Description}
Different from the single-tasking evolutionary paradigm, MFEA is a new evolutionary paradigm, considering $K$ optimization tasks concurrently, where each task contributes a factor affecting the evolution of a single population. In MFEA, a unified representation allows a unified search space made of the search spaces of all the $K$ tasks. This unified representation can be decoded into solutions of the individual tasks. The following definitions are also defined in \cite{gupta2016multifactorial} and capture the key attributes associated with each individual $ \Pi$. For simplicity, we assume all the tasks are maximization problems (see details in Section~\ref{section:prooblem Description}).

\textit{Definition 1}: The \textit{factorial cost} $f^{\Pi}$ of individual $\Pi$ measures the fitness value with respect to the $K$ tasks.

\textit{Definition 2}: The \textit{factorial rank} $r^{\Pi}_j$ of individual $\Pi$ on task  $T_j$, where $j \in \{ 1, 2, \dots, K\}$, is the index of $\Pi$ in the population sorted in descending order according to their factorial cost with respect to task $T_j$.

\textit{Definition 3}: The \textit{scalar fitness} $\varphi^{\Pi}$ of individual $\Pi$ is calculated based on its best factorial rank over the $K$ tasks, which is given by $\varphi^{\Pi} = \frac{1}{min_{j \in \{ 1, 2, \dots, K \} r^{\Pi}_j} }$.

\textit{Definition 4}: The \textit{skill factor} of individual $\Pi$  denotes the most effective task among the $K$ tasks, and is given by $\tau^{\Pi} = argmin_j \{ r^{\Pi}_j \}$, where $j \in \{ 1, 2, \dots, K\}$.

Based on the scalar fitness, evolved solutions in a population can be compared across the $K$ tasks. In particular, an individual associated with a higher scalar fitness is considered to be better. Therefore, \textit{multifactorial optimality} is defined as below:

\textit{Definition 5}: An individual $\Pi^{\star}$ associated with factorial cost $\{ f_1^{\star}, f_2^{\star}, \dots, f_K^{\star}\}$ is optimal iff $\exists j \in \{ 1, 2, \dots, K\}$  such that $f_j^{\star} \geq f(\Pi)$, where $\Pi$ denotes any solution on task $T_j$.
\subsection{Web Service Composition Problem}\label{section:prooblem Description}
In this paper, we study the semantic \textbf{W}eb \textbf{S}ervice \textbf{C}omposition problem for \textbf{M}ultiple user segments with different \textbf{Q}oSM \textbf{P}references (henceforth referred to as \textbf{WSC-MQP}). This problem has not been explicitly studied before. We perceive our problem as an evolutionary multitasking problem that aims to optimize $K$ composition tasks concurrently with respect to the $K$ user segments for better evolving high-quality solutions.

We extend the concept \emph{composition task} defined in \cite{wang2017comprehensive,wang2017gp,wang2018towards,wang2018knowledge} for supporting QoSM preferences of $K$ user segments. The preferences of one user segment is defined as an interval, such as $QoSM \in (0.75, 0.1]$. Therefore, a \emph{composition task} (also called \emph{service request}) over a given \emph{service repository} is a tuple $T_j =(I_{T}, O_{T}, cons_j)$ where $I_{T}$ is a set of task inputs, and $O_{T}$ is a set of task outputs. The inputs in $I_{T}$ and outputs in $O_{T}$ are parameters that are semantically described by concepts in a ontology $\mathcal{O}$. $cons_j$ is a QoSM preference, where $cons_j \in (QoSM_j^a, QoSM_j^b]$, $j \in \{1, 2, \dots, K \}$ and $QoSM_j^a, QoSM_j^b$ are lower and upper bounds of QoSM that are decided by data analytical techniques for each user segment.  Due to the page limit, some concepts related to the web service composition problem, such as \emph{semantic web service}, \emph{service repository}, \emph{composite service}, QoSM, and QoS are not introduced in further details in this paper, please refer to \cite{wang2017comprehensive,wang2017gp,wang2018towards,wang2018knowledge}. 

It is essential to include infeasible individuals (i.e., solutions that violate the preference of one task) into each population since infeasible composite solutions may help to find optimal solutions of other tasks. For example, one solution is infeasible for $T_1$ as it violates $cons_1$, but it is feasible to $T_2$ as it complies with $cons_2$. This solution should be included for finding optimal solutions for $T_2$. Therefore, we allow infeasible individuals in the population, but their fitness must be penalized (see details in Eq.~(\ref{eq:fitness})). According to the fitness function in Eq.~(\ref{eq:fitness}) with respect to $T_j$, we guarantee that $Fitness(\Pi)$ of an infeasible individual falls below 0.5 while $Fitness(\Pi)$ of a feasible individual falls above 0.5. Eq.~(\ref{eq:f}) measures six quality criteria in an overall quality (i.e., QoSM and QoS) for a solution $\Pi$. Eq.~(\ref{eq:qosm}) measures two quality criteria in QoSM for a solution $\Pi$. Eq.~(\ref{eq:v}) is the violations of $cons_j$ by measuring how far it is from  $QoSM(\Pi)$ in Eq.~(\ref{eq:qosm}). In particular, an infeasible individual that violates $cons_j$ more should be penalized more.

\begin{equation}
\footnotesize
\label{eq:fitness}
Fitness(\Pi) = 
\begin{cases}
	0.5 + 0.5 *  F(\Pi) & \text{ if $QoSM(\Pi) \in$ $cons_j$,}\\
	0.5 *  F(\Pi) - 0.5 * V(\Pi) & \text{ otherwise}.
\end{cases}
\end{equation}

\begin{equation}
\footnotesize
\label{eq:f}
F(\Pi) = w_1 \hat{MT} + w_2 \hat{SIM} + w_3 \hat{A} + w_4 \hat{R} + w_5(1 - \hat{T}) + w_6(1 - \hat{CT})
\end{equation}

\begin{equation}
\footnotesize
\label{eq:qosm}
QoSM (\Pi) = w_7 \hat{MT} + w_8 \hat{SIM}
\end{equation}

\begin{equation}
\footnotesize
\label{eq:v}
V(\Pi) = 
\begin{cases}
	QoSM_j^a - QoSM(\Pi) & \text{ if $QoSM(\Pi) \leq QoSM_j^a$,}\\
	QoSM(\Pi)- QoSM_j^b  & \text{ otherwise}.
\end{cases}
\end{equation}

\noindent with $\sum_{k=1}^{6} w_k= 1$ and $\sum_{k=7}^{8} w_k= 1$. We can adjust the weights according to the preferences of user segments. $\hat{MT}$, $\hat{SIM}$, $\hat{A}$, $\hat{R}$, $\hat{T}$, and $\hat{CT}$ are normalized values calculated within the range from 0 to 1 using Eq.~(\ref{eq_normal}). To simplify the presentation we also use the notation $(Q_1,Q_2,Q_3,Q_4,Q_5,Q_6) $ $= (MT,SIM,A,R,T,CT)$. $Q_1$ and $Q_2$ have minimum value 0 and maximum value 1. We refer to \cite{wang2017comprehensive,wang2017gp,wang2018knowledge} for details on the calculation of each quality criteria.

\begin{equation}
\footnotesize
\label{eq_normal}
\hat{Q_k} = 
\begin{cases}
	\frac{Q_k - Q_{k, min}}{Q_{k, max} - Q_{k, min}} & \text{ if $k=1,\ldots,4$ and }Q_{k, max} - Q_{k, min} \neq 0,\\
	\frac{Q_{k,max} - Q_k}{Q_{k, max} - Q_{k, min}} & \text{ if $k=5,6$ and }Q_{k, max} - Q_{k, min} \neq 0,\\
	1 & \text{ otherwise}.
\end{cases}
\end{equation}

\noindent To find the $K$ best possible solutions with one for each task, our goal is to maximize the objective function in Eq. (\ref{eq:fitness}) concerning the $K$ tasks.

\section{Our new method PMFEA}\label{eda_approach}
In this section, we present our new method to solve \textbf{WSC-MQP}. We begin with an overview of PMFEA, and afterwards we discuss some critical components of PMFEA in more detail.
\subsection{An overview of PMFEA}\label{subsection:outline}
 Our proposed PMFEA is characterized by three novel aspects. Firstly, we employ a permutation-based representation for composite solutions to establish a common search space over $K$ composition tasks (see details in \ref{subsection:representation}). This permutation-based representation has shown its promises in single-objective EC-based service composition approaches \cite{wang2018knowledge,da2018evolutionary}. Meanwhile, permutation-based crossover and mutation operators can be effectively used to search optimal solutions in assortative mating (see details in \ref{subsection:mating}).

Secondly, we introduce a neighborhood structure over multiple tasks for more effectively evolving solutions in PMFEA for finding high-quality solutions. By evaluating evolved solutions on neighboring tasks, we increase the chance for a solution evolved for one inherited task (determined through vertical culture transmission) to participate in building the solutions of related tasks. In our problem,  the related tasks are tasks whose QoSM preferences are adjacent to that of the inherited task. It is through this way that knowledge can be exchanged effectively across multiple tasks, enabling our algorithm to effectively cope with a problem with more than two concurrent composition requests (see details in Section~\ref{subsection:selectiveTasking}).

Thirdly, we show that fitness evaluations of a solution on neighboring tasks are fairly lightweight in \textbf{WSC-MQP} because once the calculation of $F(\Pi)$ in Eq.~(\ref{eq:fitness}) (which is very time-consuming) is completed for the inherited task and not required to be calculated again for the neighboring tasks. Moreover, we expect the execution time of PMFEA can be reduced further. Due to the effective knowledge transformation across different tasks through the introduced neighborhood structure, we increase the chances of evolving effective solutions through assortative mating. Therefore, the process of the graph-building could be accelerated by the better knowledge transformation, i.e., the order of services in a permutation for composition. In particular,  redundant services, such as $S_4$ in Fig.~\ref{fig:decoding} will not be checked before $End$ is returned (see details in Section~\ref{subsection:representation}).
\subsection{Outline of PMFEA}\label{subsection:outline}

\begin{algorithm}
 \SetKwInOut{Input}{Input}\SetKwInOut{Output}{Output}
 \SetKwFunction{generateWeightedGraph}{generateWeightedGraph}
 \SetKwProg{Procedure}{Procedure}{}{}
 \SetNlSty{}{}{:}
 \Input{$T_{j}$, $K$, and $g_{max}$}
 \Output{A set of solutions}
 Randomly initialize population $\pop{g}$ of $m$ permutations $\Pi^{g}_{k}$ as solutions (where $g=0$ and $k=1,\dots,m$)\;
 Decode each $\Pi^{g}_{k}$ into DAG $\gra^{g}_{k}$\ using a forward graph-building technique\;
 Evaluate $f^{\Pi^{g}_{k}}$,  $r^{\Pi^{g}_{k}}_j$ , $\varphi^{\Pi^{g}_{k}}$ and $\tau^{\Pi^{g}_{k}}$ of $\Pi^{g}_{k}$ over $T_{j}$, where $j \in \{1,2, \dots, K\}$\;
 \While {$g < g_{max}$}{
     Apply assortative mating to the randomly selected individuals to generate offspring population $\pop{g+1}_{a}$\;
     Assign offspring in $\pop{g+1}_{a}$ to the selected tasks and evaluate $f^{\Pi^{g+1}_{k}}$ on the tasks\;
     $\pop{g+1}$ $=$ $\pop{g}$ $\cup$ $\pop{g+1}_{a}$\;
     Update $r^{\Pi^{g+1}_{k}}_j$ , $\varphi^{\Pi^{g+1}_{k}}$ and $\tau^{\Pi^{g+1}_{k}}$ of offspring in $\pop{g+1}$\;
     Keep top half the fittest individuals in $\pop{g+1}$ based on $\varphi^{\Pi^{g+1}_{k}}$\;
 }
 Return the best $\Pi^{\star}_j$ over all the generations for $T_j$\;
\caption{PMFEA for WSC-MQP}
\label{alg:PMFEA}
\end{algorithm} 

The overview of PMFEA is summarized in \textsc{Algorithm}~\ref{alg:PMFEA}: we initially randomly generate $m$ permutations $\Pi^{g}_{k}$, where $0\leq k<m$ and $g=0$. Each permutation will be decoded into a DAG-based solution, $\gra^{g}_{k}$, see the details in Section~\ref{subsection:representation}. Subsequently, the following steps (Step 3 to 9) are repeated until a maximum generation $g_{max}$ is reached.  During the iteration,  we evaluate  $f^{\Pi^{g}_{k}}$,  $r^{\Pi^{g}_{k}}_j$ , $\varphi^{\Pi^{g}_{k}}$ and $\tau^{\Pi^{g}_{k}}$ of $\Pi^{g}_{k}$ over $T_{j}$, where $j \in \{1,2, \dots, K\}$. Afterward, we apply assortative mating to breed offspring population $\pop{g}_{a}$. In particular, crossover and mutation operators in assortative mating will be employed (see details in Section~\ref{subsection:mating}). Once $\pop{g}_{a}$ is generated, individuals in $\pop{g}_{a}$ will be assigned to tasks based on vertical cultural transmission in \textsc{Algorithm}~\ref{alg:transmission} and identified neighbouring tasks (see details in Section~\ref{subsection:selectiveTasking}) for evaluations. Consequently, we produce the next population $\pop{g+1}$ by combining the current population $\pop{g}$ and assortative mating offspring population $\pop{g}_{a}$. We update $r^{\Pi^{g}_{k}}_j$ , $\varphi^{\Pi^{g}_{k}}$ and $\tau^{\Pi^{g}_{k}}$ of the  combined population in $\pop{g+1}$, and keep the top half of fittest individuals in $\pop{g+1}$ based on $\varphi^{\Pi^{g}_{k}}$. In each generation, we keep track of the fittest $\Pi^{\star}_j$ for each task $T_j$. When the maximal generation $g_{max}$ is met,  we return the best $\Pi^{\star}_j$ over all the generations for $T_j$.
\subsection{Permutation-based representation}\label{subsection:representation}
A permutation is a sequence of all the services in the repository, and each service appears exactly once in the sequence. Each service has a unique id, i.e., index number from $0$  to $n$. Let $\Pi = (\pi_1, \dots, \pi_t, \dots, \pi_{n})$ be a permutation-based composite solution of service indexes $\{ 0, \dots, t, \dots, n\}$ such that $\pi_i \neq \pi_j$ for all $i \neq j$. Permutation-based solutions must be decoded into DAG-based solutions for easily calculating of factorial cost and presenting users a final execution services workflow \cite{wang2018knowledge}.

\begin{figure}
\footnotesize
\centering
\includegraphics[width=0.45\textwidth]{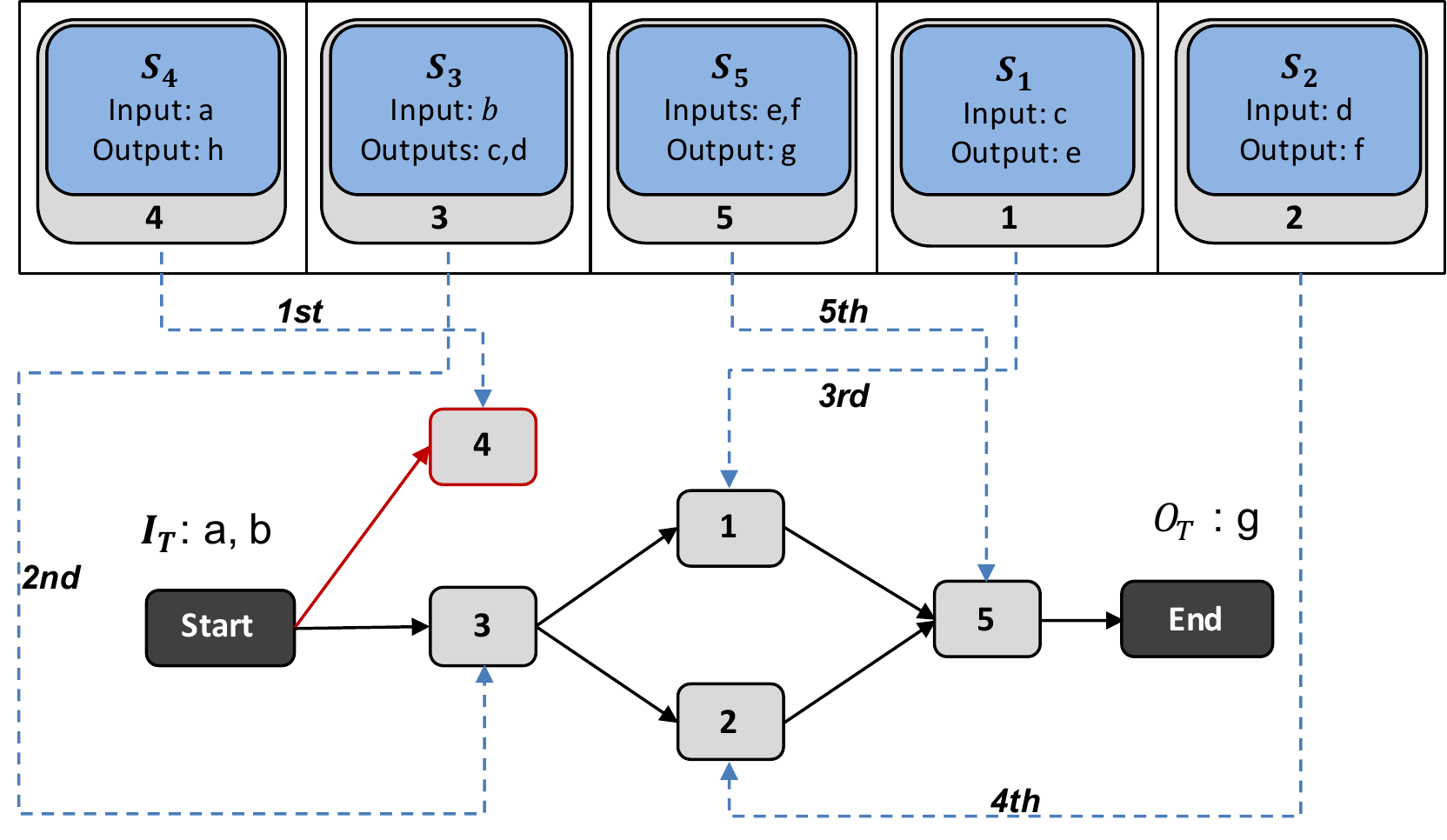}
 \caption{Example of a DAG-based solution decoded from a given permutation}
 \label{fig:decoding}
\end{figure}

Fig.~\ref{fig:decoding} illustrates an example of producing a DAG-based solution decoded from a permutation using a forward graph-building technique~\cite{wang2017comprehensive}. In the example, we take an arbitrary permutation $[4, 3, 5, 1, 2]$ as an example with composition task inputs $I_T$ and outputs $O_T$. We check the permutation from left to right, looking for services whose inputs can be fulfilled by $I_T$, so we remove them from the permutation and add them to the graph. Afterwards, we go through the permutation from left to right again and add services whose inputs can be fulfilled by $I_T$ and any outputs of services in the graph. We continue this process until we can add $End$ to the graph (i.e., $O_T$ can be produced). Note that this process may result in graphs that contain some services whose outputs are not used to fulfill the input of any other service, such as service $S_4$. These services will be removed later on.

\subsection{Assortative Mating}\label{subsection:mating}
PMFEA employs assortative mating to breeding offspring for K segment tasks. In particular, two randomly selected parent candidates undergo crossover if they have the same skill factors. Otherwise, a randomly generated probability $rand$ that is used to balance exploitation and exploration across tasks: crossover is performed over the parent candidates with different skill factors or mutation is performed on each parent, see \textsc{Algorithm}~\ref{alg:mating} in \textsc{Appendix}~\ref{appendix:mating} for technical details. 

Two-point crossover and one-point swap mutation \cite{da2018hybrid,lacomme2004competitive} for single-objective single-tasking service composition works are employed for the purpose of assortative mating to generate permutations. In a crossover, two children are produced, and each child preserves a part of the permutation from one parent while the remaining parts are filled by another parent. The mutation operator swaps the positions of two elements in the permutation. The inherited skill factors of children will be discussed in Section~\ref{subsection:selectiveTasking}

\begin{figure}
\footnotesize
\centering
\includegraphics[width=0.485\textwidth]{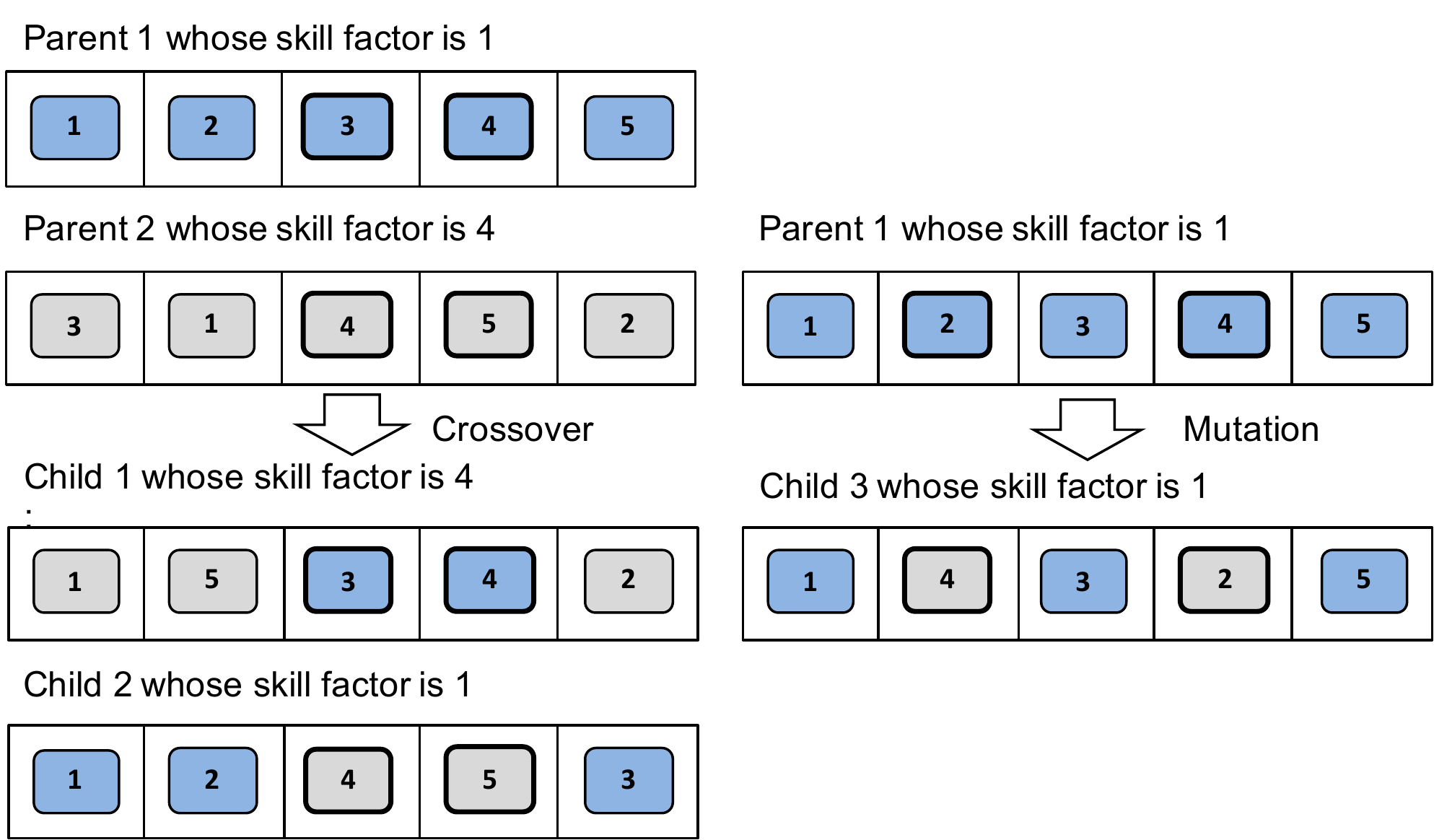}
 \caption{Examples of crossover and mutation for two parents with different skill factors}
 \label{fig:crossover}
\end{figure}

Fig.~\ref{fig:crossover} illustrates an example of crossover and mutation for randomly selected parents with different skill factors, e.g., the skill factors of the platinum and the bronze segment are 1 and 4  respectively. In a  crossover, Child 1 preserves positions of 3 and 4 of Parent 1 while the other parts are filled from left to right with 1, 5, and 2 that are obtained from Parent 2 from its left to right. Child 2 is also produced in the same way. In a mutation, Child 3 is produced by swapping the positions of 2 and 4 in Parent 1.

\subsection{Task Selection for Evaluations}\label{subsection:selectiveTasking}
Gupta et al.~\cite{gupta2016multifactorial} investigated two optimization tasks and suggest candidate solutions is only evaluated on the task that is inherited from parents based on the vertical cultural transmission, see details in \textsc{Algorithm}~\ref{alg:transmission}  in \textsc{Appendix}~\ref{appendix:transmission} for details. 

To effectively deal with more than two optimization tasks, we proposed PMFEA-NT for evolving more effective solutions through careful selections of tasks, where we introduce a neighborhood structure over a set of tasks. In particular, we suggest identifying the neighboring tasks of each child's inherited task, which is imitated from the vertical culture transmission. Subsequently, we will assign each child to the neighboring tasks for additional evaluations.

\begin{figure}
\footnotesize
\centering
\includegraphics[width=0.38\textwidth]{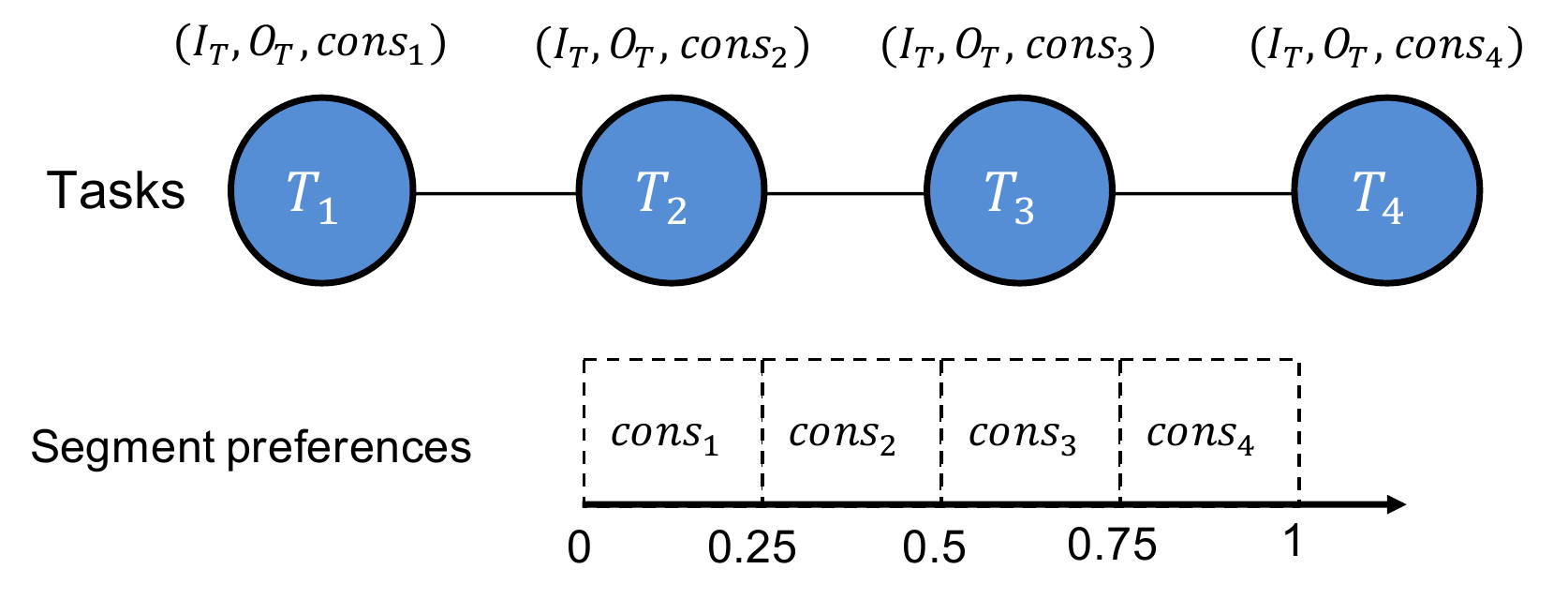}
 \caption{Examples of neighborhood structure over four tasks}
 \label{fig:neighbor}
\end{figure}
\begin{table*}
\scriptsize
\centering
\caption{Mean fitness values for our approach in comparison to FL~\cite{da2018evolutionary}\\ (Note: the higher the fitness the better)}
\label{tbl:meanFitness}
\begin{tabular}{|c||c|c|c||c|}
  \hline
   \multicolumn{5}{|c|}{Task 1} \\
   \hline\hline
  \rule[2mm]{0pt}{0mm}
     Method & PMFEA-AT        & PMFEA-NT               & PMFEA             &FL~\cite{da2018evolutionary}\\ 
  \hline\hline
        WSC09-1 & $\mathbf{0.192631} \pm \mathbf{0.00475}$ & $\mathbf{0.19291} \pm \mathbf{0.003977}$ & $\mathbf{0.192864} \pm \mathbf{0.003543}$ & $\mathbf{0.193947} \pm \mathbf{0.003276}$ \\ 
        WSC09-2 & $\textloss{0.146518} \pm \textloss{0.003685}$ & $\mathbf{0.146975} \pm \mathbf{0.005269}$ & $\mathbf{0.148709} \pm \mathbf{0.00488}$ & $\textloss{0.146254} \pm \textloss{0.002946}$ \\ 
        WSC09-3 & $\mathbf{0.152277} \pm \mathbf{0.003385}$ & $\mathbf{0.152809} \pm \mathbf{0.003959}$ & $\textloss{0.150053} \pm \textloss{0.003885}$ & $\mathbf{0.15355} \pm \mathbf{0.002688}$ \\ 
        WSC09-4 & $\mathbf{0.141319} \pm \mathbf{0.000747}$ & $0.140892 \pm 0.000836$ & $\textloss{0.140255} \pm \textloss{0.00073}$ & $\mathbf{0.141451} \pm \mathbf{0.000515}$ \\ 
        WSC09-5 & $\mathbf{0.144593} \pm \mathbf{0.001096}$ & $\mathbf{0.144193} \pm \mathbf{0.00102}$ & $\textloss{0.143447} \pm \textloss{0.000946}$ & $\mathbf{0.144942} \pm \mathbf{0.000705}$ \\ 
   \hline\hline
   \multicolumn{5}{|c|}{Task 2} \\
   \hline\hline
  \rule[2mm]{0pt}{0mm}
     Method & PMFEA-AT        & PMFEA-NT               & PMFEA             &FL~\cite{da2018evolutionary}\\ 
       \hline\hline
        WSC09-1 & $\mathbf{0.810555} \pm \mathbf{0.00638}$ & $\mathbf{0.808541} \pm \mathbf{0.006982}$ & $\mathbf{0.809369} \pm \mathbf{0.007696}$ & $\textloss{0.807483} \pm \textloss{0.005398}$ \\ 
        WSC09-2 & $\textloss{0.748537} \pm \textloss{0.006183}$ & $\mathbf{0.749583} \pm \mathbf{0.00816}$ & $\mathbf{0.752848} \pm \mathbf{0.006956}$ & $\textloss{0.74895} \pm \textloss{0.006425}$ \\ 
        WSC09-3 & $\mathbf{0.765014} \pm \mathbf{0.007071}$ & $\mathbf{0.764333} \pm \mathbf{0.007089}$ & $\textloss{0.760746} \pm \textloss{0.006192}$ & $\textloss{0.761924} \pm \textloss{0.006194}$ \\ 
        WSC09-4 & $\mathbf{0.739807} \pm \mathbf{0.000696}$ & $\mathbf{0.73975} \pm \mathbf{0.000825}$ & $\mathbf{0.739866} \pm \mathbf{0.000853}$ & $\mathbf{0.739826} \pm \mathbf{0.000692}$ \\ 
        WSC09-5 & $\mathbf{0.73927} \pm \mathbf{0.00081}$ & $\mathbf{0.739328} \pm \mathbf{0.00073}$ & $\mathbf{0.73936} \pm \mathbf{0.001217}$ & $\mathbf{0.739467} \pm \mathbf{0.000735}$ \\    \hline\hline
   \multicolumn{5}{|c|}{Task 3} \\
   \hline\hline
  \rule[2mm]{0pt}{0mm}
     Method & PMFEA-AT        & PMFEA-NT               & PMFEA             &FL~\cite{da2018evolutionary}\\ 
       \hline\hline
        WSC09-1 & $\mathbf{0.820082} \pm \mathbf{0.00571}$ & $\mathbf{0.820097} \pm \mathbf{0.004829}$ & $\mathbf{0.820107} \pm \mathbf{0.007241}$ & $\mathbf{0.819418} \pm \mathbf{0.003768}$ \\ 
        WSC09-2 & $\textloss{0.230114} \pm \textloss{0.006103}$ & $\mathbf{0.231639} \pm \mathbf{0.007691}$ & $\mathbf{0.234557} \pm \mathbf{0.00651}$ & $\textloss{0.22968} \pm \textloss{0.004616}$ \\ 
        WSC09-3 & $\mathbf{0.788258} \pm \mathbf{0.003952}$ & $\mathbf{0.788829} \pm \mathbf{0.00307}$ & $\mathbf{0.789012} \pm \mathbf{0.002968}$ & $\mathbf{0.788726} \pm \mathbf{0.002576}$ \\ 
        WSC09-4 & $\mathbf{0.224035} \pm \mathbf{0.001628}$ & $\mathbf{0.224026} \pm \mathbf{0.001827}$ & $\mathbf{0.224278} \pm \mathbf{0.001957}$ & $\mathbf{0.224127} \pm \mathbf{0.001467}$ \\ 
        WSC09-5 & $\mathbf{0.221114} \pm \mathbf{0.001512}$ & $\mathbf{0.221319} \pm \mathbf{0.001286}$ & $\mathbf{0.221169} \pm \mathbf{0.002244}$ & $\mathbf{0.221102} \pm \mathbf{0.001248}$ \\     \hline\hline
   \multicolumn{5}{|c|}{Task 4} \\
   \hline\hline
  \rule[2mm]{0pt}{0mm}
     Method & PMFEA-AT        & PMFEA-NT               & PMFEA             &FL~\cite{da2018evolutionary}\\ 
       \hline\hline
        WSC09-1 & $\mathbf{0.222976} \pm \mathbf{0.007486}$ & $\mathbf{0.223659} \pm \mathbf{0.008279}$ & $\mathbf{0.219863} \pm \mathbf{0.013342}$ & $\mathbf{0.221582} \pm \mathbf{0.00946}$ \\ 
        WSC09-2 & $\textloss{0.105114} \pm \textloss{0.006103}$ & $\mathbf{0.10656} \pm \mathbf{0.007747}$ & $\mathbf{0.109708} \pm \mathbf{0.0   0659}$ & $\textloss{0.10468} \pm \textloss{0.004616}$ \\ 
        WSC09-3 & $\mathbf{0.215947} \pm \mathbf{0.00718}$ & $\mathbf{0.215877} \pm \mathbf{0.007496}$ & $\mathbf{0.217783} \pm \mathbf{0.005575}$ & $\mathbf{0.216698} \pm \mathbf{0.00533}$ \\ 
        WSC09-4 & $\mathbf{0.099035} \pm \mathbf{0.001628}$ & $\mathbf{0.098941} \pm \mathbf{0.001889}$ & $\mathbf{0.099276} \pm \mathbf{0.001935}$ & $\mathbf{0.099127} \pm \mathbf{0.001467}$ \\ 
        WSC09-5 & $\mathbf{0.096114} \pm \mathbf{0.001512}$ & $\mathbf{0.096312} \pm \mathbf{0.001287}$ & $\mathbf{0.096085} \pm \mathbf{0.002181}$ & $\mathbf{0.096102} \pm \mathbf{0.001248}$ \\     \hline
\end{tabular}
\end{table*}

Fig.~\ref{fig:neighbor} illustrates an example of neighborhood structure over four composition tasks $T_1$, $T_2$, $T_3$ and $T_4$  with respect to four user segments. These four composition tasks have the same input and output (i.e., $I_T$ and $I_O$) but different $cons_j$, where $j \in \{1, 2, 3, 4\}$. In particular, $cons_1 \in (0, 0.25]$, $cons_2 \in (0.25, 0.5]$, $cons_3 \in (0.5, 0.75]$ and $cons_4 \in (0.75, 1]$, respectively.  The neighborhood structure is determined based on the tasks whose segment preferences on QoSM are adjacent to each other. For example, the neighboring tasks of task $T_2$ are $T_1$ and $T_3$ whose segment preference on QoSM (i.e.,  $cons_1$ and  $cons_3$) are adjacent to that of $T_2$ (i.e., $cons_2$).  

We continue to use the example in Fig.~\ref{fig:neighbor} to demonstrate the benefits of our proposed neighborhood structure in PMFEA-NT. Consider a child derived from a parent that satisfies $cons_1$ of $T_1$, and this child can also lead to the satisfaction of a neighboring segment preference, i.e., $cons_2$. If this child is only evaluated on task $T_1$ based on the vertical cultural transmission, resulting in a poor fitness value, it is likely to be discarded. On the other hand, if we give this child a chance to be evaluated on the neighboring task, i.e.,  $T_2$, resulting in a good fitness value, it can survive to the next generation due to its good performance on $T_2$. We hope such a situation will help to diversify solutions in the population and make the evolution process more efficient.
\section{Experimental Evaluation}\label{section:experiments}

We conduct experiments to evaluate the effectiveness and efficiency of PMFEA, PMFEA-NT, and PMFEA-AT. These three approaches are compared to one state-of-art single-tasking EC-based method, i.e., Fixed Length Genetic Algorithm  (FL)~\cite{da2018evolutionary}, which is reported as a very effective method to find high-quality solutions for single-tasking service composition problem. In particular, these PMFEA approaches are utilized to find optimal solutions for Task 1, Task 2, Task 3 and Task 4 concurrently while FL is utilized to optimize each task one by one. One benchmark, i.e., WSC-09 \cite{kona2009wsc} extended with QoS attributes from~\cite{al2007qos} is used as a benchmark, which is popularly used in related works \cite{wang2017comprehensive,yu2013adaptive}. This benchmark contains five datasets, i.e., WSC09-1 to WSC09-5. Each dataset that includes one $I_{T}$, one $O_{O}$ is extended with four pre-defined QoSM preferences of user segments: $(0, 0.25]$, $(0.25, 0.5]$, $(0.5, 0.75]$, and $(0.75, 1]$. Furthermore, to demonstrate that PMFEA can maintain high performance on large-scale problems, we double the size of the service repository for each task with 1144, 8258, 16276, 16602, and 30422 services respectively.

The size of the population $m$ is set to 30, which strictly follow the population size of FL~\cite{da2018evolutionary}. The assortative mating $rand$ is set to 0.3,  following the popular evolutionary multitasking setting in \cite{yuan2016evolutionary}. The maximum generation $g_{max}$ is 200. For the compared $FL$ \cite{da2018evolutionary}, we use its reported setting: crossover and mutation are 0.95 and 0.05 respectively, tournament size is set to 2 and elitism is set to 2. The weights in the fitness function Eq.~(\ref{eq:f}) are set to balance quality criteria in both QoSM and QoS, i.e., $w_1$ and $w_2$ are set to 0.25, and $w_3$, $w_4$, $w_5$ and $w_6$ to 0.125 \cite{wang2018knowledge}. The weights Eq.~(\ref{eq:qosm})  are set to balance all quality criteria in QoSM, i.e., $w_7$ and $w_8$ are set to 0.5. We have also conducted tests with other weights and parameters and generally observed the same behavior.

\subsection{Comparison of the Fitness}\label{subsection:results}

\begin{table*}
\scriptsize
\centering
\caption{Mean execution time (in s) for our approach in comparison to \cite{da2018evolutionary}\\ (Note: the shorter the time the better)}
\label{tbl:meanTime}
\begin{tabular}{|c||c|c|c||c|}
  \hline
   \multicolumn{5}{|c|}{All tasks (Task 1, Task 2, Task 3 and Task 4)} \\
   \hline\hline
  \rule[2mm]{0pt}{0mm}
     Method & MFEA-AT        & MFEA-NT               & MFEA             &FL~\cite{da2018evolutionary}\\ 
  \hline\hline
        WSC09-1 & $\mathbf{54} \pm \mathbf{52}$ & $\mathbf{44} \pm \mathbf{32}$ & $79 \pm 87$ & $\textloss{150} \pm \textloss{151}$ \\ 
        WSC09-2 & $\mathbf{1900} \pm \mathbf{1032}$ & $\mathbf{1925} \pm \mathbf{702}$ & $2371 \pm 804$ & $\textloss{8479} \pm \textloss{3002}$ \\ 
        WSC09-3 & $\mathbf{1479} \pm \mathbf{1257}$ & $\mathbf{1542} \pm \mathbf{1159}$ & $\mathbf{1821} \pm \mathbf{740}$ & $\textloss{5926} \pm \textloss{3199}$ \\ 
        WSC09-4 & $64311 \pm 16843$ & $\mathbf{60925} \pm \mathbf{16311}$ & $71903 \pm 19042$ & $\textloss{250146} \pm \textloss{55355}$ \\ 
        WSC09-5 & $\mathbf{12943} \pm \mathbf{6615}$ & $\mathbf{12456} \pm \mathbf{6094}$ & $\mathbf{13689} \pm \mathbf{6723}$ & $\textloss{47879} \pm \textloss{16126}$ \\ 
  \hline
\end{tabular}
\end{table*}

We use an independent-sample T-test with a significance level of 5\% to verify the observed differences in mean fitness over 30 runs.  In particular, a pairwise comparison of approaches was carried to rank the performances of all the approaches based on the number of times they were found to be better, similar, or worse than the others. We highlight the best performances and the worst performances in   green and red colors, respectively, for related values in the tables. Note that all values in a row are highlighted in green implying no significant differences among all the approaches for the task.

First, all the multitasking approaches, i.e., PMFEA, PMFEA-NT, and PMFEA-AT, outperform FL~\cite{da2018evolutionary} since the most values related to FL~\cite{da2018evolutionary} in Tables~\ref{tbl:meanFitness} are marked in red color. This observation agrees with the findings in work \cite{gupta2016multifactorial} that multitasking is more competent at improving the quality of solutions by utilizing the knowledge of other tasks through assortative mating.

Second, the quality of solutions produced by MFEA-NT is the most favorable one since all the solutions are marked as best performance except one that is marked with an average performance in Tables~\ref{tbl:meanFitness}. This corresponds well with our expectation that careful selections of the neighborhood structure can contribute to searching good solutions effectively.  It is due to that the implicit knowledge of solutions (i.e., the order of services used for composition) on one task can potentially be more effectively transferred to the neighboring tasks. 

Third, MEFA  and MFEA-AT are comparable to each other, and both are less favorable to MFEA-NT since its performance varies on different tasks, i.e., 16 out of 20 tasks as best performance and 4 out of 20 tasks are marked as worst performance in Tables~\ref{tbl:meanFitness}. MEFA strictly follows the vertical cultural transmission, so it loses the chance to transfer implicit knowledge to other tasks. On the other hand, although MFEA-AT assigns candidate solutions to all tasks for evaluations, it can be easily trapped in local optima, e.g., all the four tasks in WSC09-2.  It may due to that candidate solutions only inherit the most effective task over all the tasks and make a locally optimal choice each time. Such a greedy strategy can easily lead to local optima.
\subsection{Comparison of the Execution Time}\label{subsection:results}
Table~\ref{tbl:meanTime} shows the execution times observed for MFEA-AT, MFEA-NT, MFEA and FL~\cite{da2018evolutionary} on the four tasks as a whole. Again an independent-samples T-test has been conducted over 30 runs.  

MFEA, MFEA-NT, and MFEA-AT require significantly less execution time while FL~\cite{da2018evolutionary} consistently takes four times the execution time of MFEA, MFEA-NT and MFEA-AT approximately in the worst cases. e.g., mean execution time for WSC09-3. It is due to that FL~\cite{da2018evolutionary} is a single-tasking EC technique that optimizes four tasks separately, unlike MFEA, MFEA-NT and MFEA-AT.

MFEA-NT achieves the shortest execution time for each dataset consistently. Meanwhile, MFEA-NT and MFEA-AT are very comparable to each other. It corresponds well with our expectation that the execution time of PMFEA can be reduced due to the effective knowledge transformation across different tasks through the introduced neighborhood structure over multiple tasks.
\subsection{Comparison of the Convergence Rate }\label{subsection:convergence}
\begin{figure}
\label{fig:convergence}
	\caption{Mean fitness over generations for tasks 1-4, for WSC09-3 (Note: the larger the fitness the better)}
    \centering
	\begin{subfigure}{0.43\columnwidth} 
		\includegraphics[width=\textwidth]{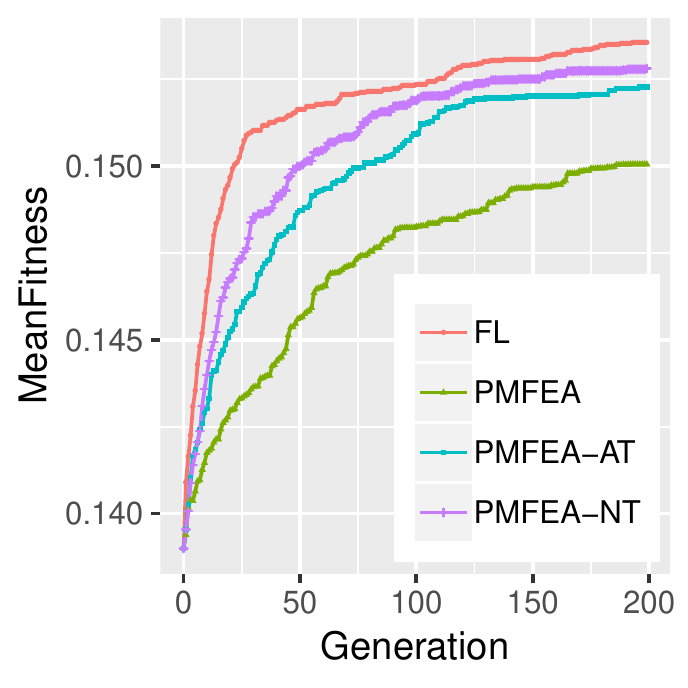}
		\caption{Task 1} 
	\end{subfigure}
	\begin{subfigure}{0.43\columnwidth} 
		\includegraphics[width=\textwidth]{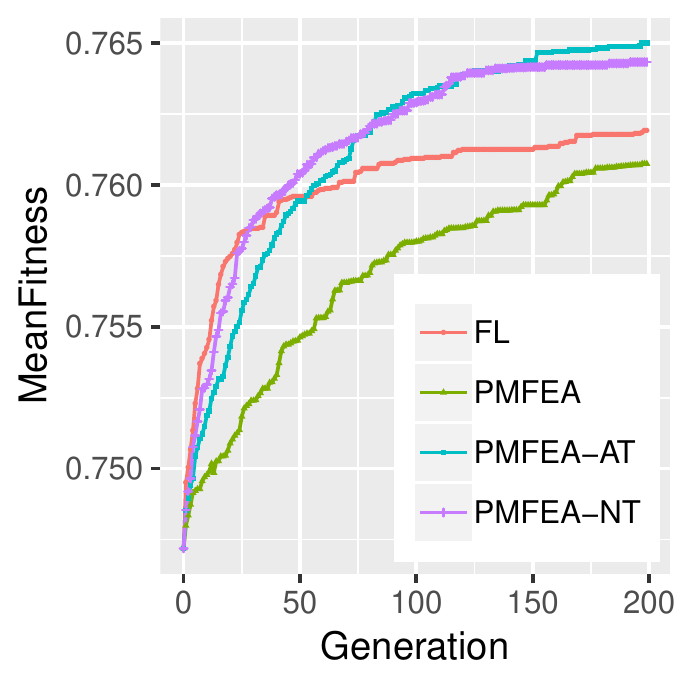}
		\caption{Task 2} 
	\end{subfigure}
		\begin{subfigure}{0.43\columnwidth} 
		\includegraphics[width=\textwidth]{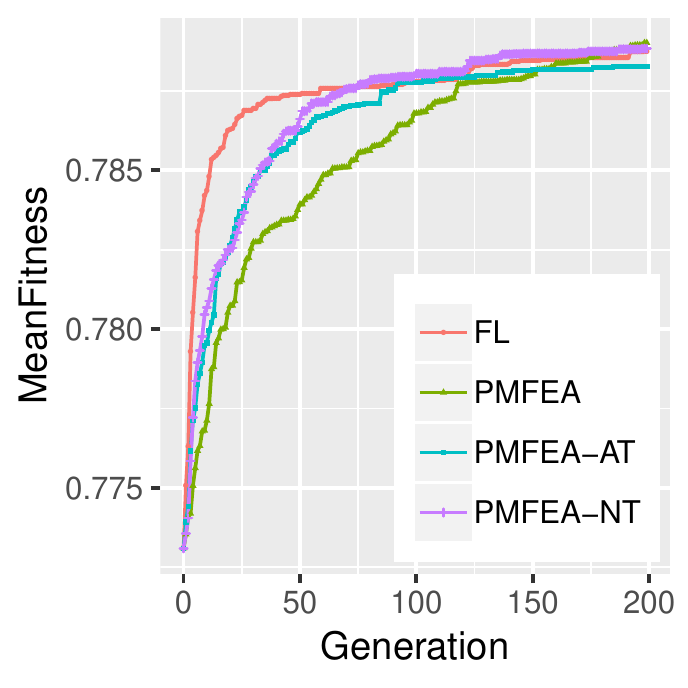}
		\caption{Task 3} 
	\end{subfigure}
	\begin{subfigure}{0.43\columnwidth} 
		\includegraphics[width=\textwidth]{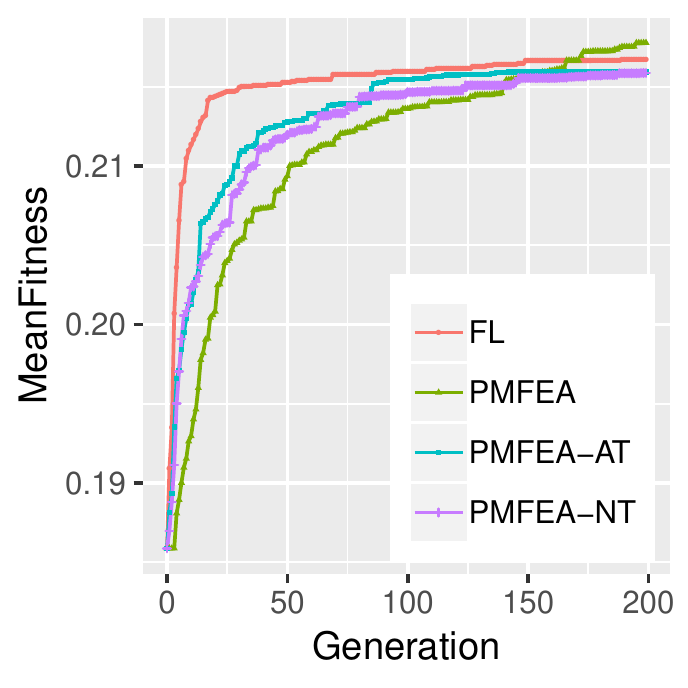}
		\caption{Task 4} 
	\end{subfigure}
\end{figure}
We investigate the convergence rate of PMFEA-AT, PMFEA-NT, PMFEA, and FL~\cite{da2018evolutionary} on four tasks over 30 runs, and use WSC09-3 as an example to illustrate the performance of all the compared methods.

Fig.~4 shows the evolution of the mean fitness value of the best solutions found along 200 generations for all the approaches. Among all the four tasks, we observe a significant increase in the fitness value towards the optimum for all the approaches, which eventually reach a plateau with more stable improvements. In particular, PMFEA converges much slower consistently than all the other approaches. PMFEA-NT and PMFEA-AT converge much faster than PMFEA and are comparable to each other. This observation further agrees with our findings that evaluating an offspring on neighbor tasks or all tasks are essential for multitasking service composition with more than two concurrent composition tasks. On the other hand, FL~\cite{da2018evolutionary}  happens to converge very fast at early generations, but PMFEA-AT, PMFEA-NT eventually get a chance to catch up with FL~\cite{da2018evolutionary} in later generations, such as convergence rates over Task 1, 3 and 4. 
\section{Conclusion}\label{conclusion}
In this paper, we model multiple service composition tasks for user segments with different QoSM preferences as a multi-tasking problem and propose a permutation-based multifactorial evolutionary algorithm to solve this problem. We also introduce a neighborhood structure over multiple tasks to allow newly evolved solutions to be evaluated on related tasks without incurring extra computation time. This structure is vital for supporting more than two tasks. Our proposed method can perform better at the cost of only a fraction of time, compared to one state-of-art single-tasking EC-based method. We also found that the use of the proper neighborhood structure can enhance the effectiveness of our approach.
\bibliographystyle{IEEEtranBib}
\small{
\bibliography{IEEEexample} 
}
\appendix
\subsection{Assortative Mating}\label{appendix:mating}
The procedure of assortative mating for breeding offspring for K composition tasks is outlined in \textsc{Algorithm}~\ref{alg:mating}.
\begin{algorithm}
 \SetKwInOut{Input}{Input}\SetKwInOut{Output}{Output}
 \SetKwFunction{generateWeightedGraph}{generateWeightedGraph}
 \SetKwProg{Procedure}{Procedure}{}{}
 \SetNlSty{}{}{:}
 Randomly select two parents $\Pi^{g}_a$ and $\Pi^{g}_b$ from $\pop{g}$\;
 $rand$ $\leftarrow$ $Rand(0,1)$\;
 \eIf{$\tau^{\Pi^{g}_a} = \tau^{\Pi^{g}_b}$ or $rand < rmp$}{    
        Perform crossover on $\Pi^{g}_a$ and $\Pi^{g}_b$ to generate two children $\Pi^{g}_c$ and $\Pi^{g}_d$\;
       }{
        Perform mutation on $\Pi^{g}_a$ to generate one child $\Pi^{g}_e$\;
        Perform mutation on $\Pi^{g}_b$ to generate one child $\Pi^{g}_f$\;
       }
\caption{Assortative Mating \cite{gupta2016multifactorial}}
\label{alg:mating}
\end{algorithm} 
\subsection{Vertical Cultural Transmission}\label{appendix:transmission}
Vertical cultural transmission via selective imitation is illustrated in \textsc{Algorithm}~\ref{alg:transmission}, where any child produced by assortative mating is only evaluated on one selected task that is determined by the skill factors of its parents.
\begin{algorithm}
 \SetKwInOut{Input}{Input}\SetKwInOut{Output}{Output}
 \SetKwFunction{generateWeightedGraph}{generateWeightedGraph}
 \SetKwProg{Procedure}{Procedure}{}{}
 \SetNlSty{}{}{:}
 \eIf{$\Pi^{g}_k$ is produced by two parents $\Pi^{g}_a$ and $\Pi^{g}_b$}{
        Generate a random $rand$ between 0 and 1\;
        \eIf{$rand < 0.5$}{
        $\Pi^{g}_k$ imitates the skill factor $\tau^{\Pi^{g}_a}$ of $\Pi^{g}_a$\;
        $\Pi^{g}_k$ is only evaluated on task $T_{\tau^{\Pi^{g}_a}}$\;
        }{
        $\Pi^{g}_k$ imitates the skill factor $\tau^{\Pi^{g}_b}$ of $\Pi^{g}_b$\;
        $\Pi^{g}_k$ is only evaluated on task $T_{\tau^{\Pi^{g}_b}}$\;

        }
       }{
        Let $\Pi^{g}_e$ be the only one parent of $\Pi^{g}_k$\;
        $\Pi^{g}_k$ imitates the skill factor $\tau^{\Pi^{g}_e}$ of $\Pi^{g}_e$\;
        $\Pi^{g}_k$ is only evaluated on task $T_{\tau^{\Pi^{g}_e}}$\;
       }
\caption{Vertical Cultural Transmission Via Selective Imitation \cite{gupta2016multifactorial}}
\label{alg:transmission}
\end{algorithm} 
\end{document}